# Multi-Sensor Fusion Method using Dynamic Bayesian Network for Precise Vehicle Localization and Road Matching

Cherif Smaili[1], Maan E. El Najjar[2], François Charpillet[1]

[1]*LORIA-INRIA Lorraine - MAIA Team Campus Scientifique BP 239 54506 Vandoeuvre-lés-Nancy, France*

[2]*LAGIS-CNRS UMR 8146 Polytech'Lille, Avenue Paul Langevin 59655 Villeneuve d'Ascq Cedex France*

smailic@loria.fr

**Abstract**

*This paper presents a multi-sensor fusion strategy for a novel road-matching method designed to support real-time navigational features within advanced driving-assistance systems. Managing multi-hypotheses is a useful strategy for the road-matching problem. The multi-sensor fusion and multi-modal estimation are realized using Dynamical Bayesian Network. Experimental results, using data from Anti-lock Braking System (ABS) sensors, a differential Global Positioning System (GPS) receiver and an accurate digital roadmap, illustrate the performances of this approach, especially in ambiguous situations.*

## 1. Introduction

Autonomous Vehicles currently hold the attention of many researchers because they can provide solutions in many applications related to Intelligent Transportation system. One example of such a system is the transport of passengers in urban environments using a CyCab [1]. For navigational needs the vehicle first needs to know its position on the road network, and then to retrieve attributes from the appropriate databases. Examples of attributes are maximum authorized speed, the width of the road, the presence of landmarks for precise localization, etc. Unfortunately, the precise localization on a digital map cannot be guaranteed because there will often be errors in the estimation of position arising from sensor imprecision and because the map represents a deformed view of the real world: roads are represented by points – nodes and shaping points- that describe the geometry of the center line.

Vehicle localization on a map has two meanings in the literature in this domain. In many works, [2], [3], [4] and [5] it refers to the projection of the absolute position estimate onto a segment of the road network stored in the database. In this case, the vehicle is localized when the curvilinear abscissa along the segment are known from the starting node. These "arc-matching" methods therefore introduce geometric distortions, since the model of the world is a set of segments, usually with a 10 meters absolute error and a 1 meter relative error. Alternatively, vehicle localization can refer to absolute localization in the map reference frame. In this case, the localization of the vehicle does not need a projection onto the segments representing the road in the database. Absolute localization can be very useful for the following reasons. In several kinds of databases, including those of the National French Institute of Geography (IGN), attributes, instead of being attached to the arcs representing the roads, can be stored in the database as point objects with an absolute position.

The approach presented in this paper is an absolute localization method. The global positions provided by a GPS receiver are projected onto the map frame. The goal is to select the most likely segment(s) from a set of segments close to the estimation of the vehicle position. Nowadays, since the geometry of roadmaps is more and more detailed, the number of segments representing roads is increasing. The road managing module is an important stage in the vehicle localization process because the robustness of the localization depends mainly on this stage. In order to focus in this point, an accurate map Géoroute V2 provided by the IGN was used in this work.

In order to develop our approach, it is important to estimate continuously the pose – position and heading – of the vehicle in the frame of the map using GPS, because of its affordability and convenience. However, GPS suffers from satellite outages occurring in urban environments, under bridges, tunnels or in forests. GPS can thus be seen as an intermittently-available positioning system that needs to be backed up by a dead-reckoning system [7]. In this work, a low-cost odometric method based on the use of encoders attached to the rear wheels is proposed. A dead-reckoned estimated pose is obtained by integrating the elementary rotations of the wheels starting from a known pose. The multisensor fusion of GPS and odometry is performed by a Switching Kalman Filter (SKF). This kind of formalism is also useful in quantifying the imprecision associated with each estimated pose.

The outline is as follows. In section 2, the architecture of the road-matching method is described. In section 3, we present a short introduction of Bayesian networks. Section 4 describes the network model for road-matching. Next section, we evaluate the potential of our approach by presenting a real data results. Finally, concluding remarks are given in section 6.

## 2. Architecture of the road-matching strategy

The road-matching method described in this section relies on Bayesian networks. The proposed approach is described in Figure 1. Firstly, the algorithm combines the ABS measurements with a GPS position, if it is available. Then, using this estimate, segments around the estimation are selected in a radius of 30 meters by using a Geographical Information System (GIS)-2D. Using these segments, map observations are built and merged with other data sensors using a method based on Dynamic Bayesian Network.

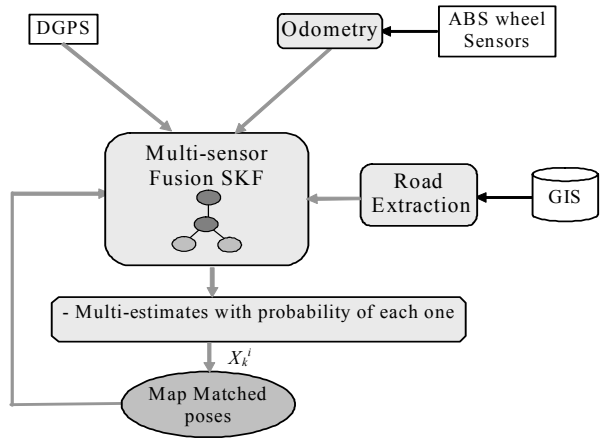

**Figure 1: Synoptic of the road-matching method**

### 2.1 Localization and heading estimation by combining odometry and GPS

Consider a car-like vehicle with front-wheel drive. The mobile frame is chosen with its origin M attached to the center of the rear axle. The x-axis is aligned with the longitudinal axis of the car (see Figure 2).

The vehicle's position is represented by the $(x_k, y_k)$ Cartesian coordinates of M in a world frame. The heading angle is denoted $\theta_k$. If the road is perfectly planar and horizontal, and if the motion is locally circular, the motion model can be expressed as [8]:

$$X_k = \begin{cases} x_{k+1} = x_k + \delta_s \cdot \cos(\theta_k + \delta_\theta/2) \\ y_{k+1} = y_k + \delta_s \cdot \sin(\theta_k + \delta_\theta/2) \\ \theta_{k+1} = \theta_k + \delta_\theta \end{cases} \quad (1)$$

Where $\delta_s$ is the length of the circular arc followed by M and $\delta_\theta$ is the elementary rotation of the mobile frame. These values are computed using the ABS measurements of the rear wheels. Let denote $X_k$ the state vector containing the pose.

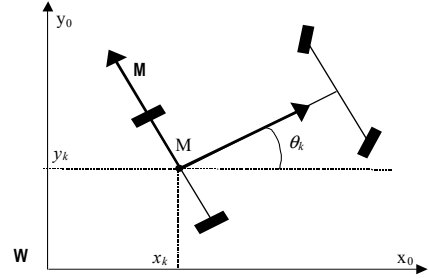

**Figure 2: The mobile frame attached to the car**

### 2.2 Cartographical and GPS observation equation

A correction of the odometric estimation is performed by the GPS information. When the GPS satellites signal is blocked by buildings or tunnels, the odometric estimation is used to select the segments all around the estimation from the cartographical database. The cartographical observations can be obtained by projections onto the segments. If the orthogonal projection onto line does not make part of the segment, the closer extremity is used (see Figure 3). When several segments are candidates, the cartographical observation function is a non-linear multi-modal observation. Considering a Gaussian distribution of noise to represent the uncertainty zone all around a segment, so the multi-modal observation is a multi-Gaussian observation one.

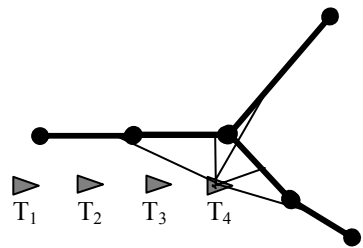

**Figure 3: Most likely segments extracted from the database**

The observation equation of the segment $seg_i$ can be written:

$$Y_{carto}^{segi} = \begin{bmatrix} x_{carto} \\ y_{carto} \\ cap_{carto} \end{bmatrix} = \begin{bmatrix} 1 & 0 & 0 \\ 0 & 1 & 0 \\ 0 & 0 & 1 \end{bmatrix} \cdot \begin{pmatrix} x \\ y \\ \theta \end{pmatrix} + \beta_{carto}$$

Where ($x_{carto}$, $y_{carto}$) is the orthogonal projection onto each segments and $cap_{carto}$ is the segment heading.

To represent the error of the cartographical observation in the SKF formalism, we choose a Gaussian distribution of the uncertainty zone all around the segment. So this error can be represented with an ellipsoid which encloses the road (we choose to use an ellipsoid because it is just the available model). This ellipsoid has its semi-major axis in the length of the segment and its semi-minor axis equals to the width of the road [8] (see Figure 4).

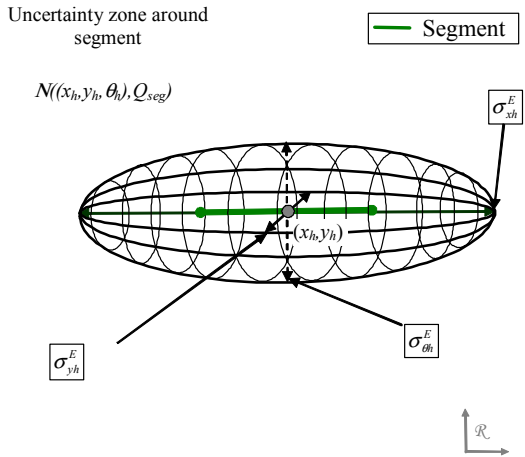

**Figure 4: Ellipsoidal of probability construction representing zone around a segment for horizontal segment i.e. parallel with the east axes**

The third axis of the ellipsoid represents the uncertainty of the estimation of the segment. This uncertainty is related to the relative error of the cartographical database. The covariance matrix of the cartographical observation error can be written:

$$Q_k^{carto} = \begin{pmatrix} \sigma_{x,h}^2 & Q_{xy,h} & 0 \\ Q_{xy,h} & \sigma_{y,h}^2 & 0 \\ 0 & 0 & \sigma_{\theta,h}^2 \end{pmatrix}$$

The GPS position measurement provides the GPS observation ($x_{gps}$, $y_{gps}$). The GPS measurement error can be provided also and in real time using the Standard National Marine Electronics Association (NMEA) sentence "GPGST" given by the Trimble AgGPS132 receiver which has been used in the experiments. Therefore, the GPS noise is not stationary. The non stationary of the GPS measurements noise affect the observation model. With each measurement provided, the GPS provide it with his noise in the sequence GPGST in the standard NMEA.

$Q_k^{gps}$ : covariance matrix of the GPS error where

$$Q_k^{gps} = \begin{pmatrix} \sigma_{x,gps}^2 & Q_{xy,gps} \\ Q_{xy,gps} & \sigma_{y,gps}^2 \end{pmatrix}_k$$

The observation equation can be written:

$$Y_1 = \begin{bmatrix} x_{gps} \\ y_{gps} \end{bmatrix} = \begin{bmatrix} 1 & 0 & 0 \\ 0 & 1 & 0 \end{bmatrix} \cdot \begin{pmatrix} x \\ y \\ \theta \end{pmatrix} + \beta_{gps}$$

## 3. Bayesian Networks

A Bayesian Network (BN) is a graph with probabilities for representing random variables and their dependencies. It efficiently encodes the joint probability distribution (JPD) of a set of variables. Its nodes represent random variables and its arcs represent dependencies between the random variables encoded by conditional probabilities. Afterwards, a BN is written as a directed acyclic graph G=(E, W) with W={$X_1$,…, $X_N$} as the set of nodes, and ($X_i$,$X_j$)∈ E, the set of edges, if $X_i$ ∈ Pa($X_j$). Normally an edge is drawn from $X_i$ to $X_j$ if $X_i$ has a direct influence on $X_j$. For more detailed in Bayesian Networks see [9] [10].

The joint probability distribution of random variables S= {$X_1$,…,$X_N$} in a BN is calculated by the multiplication of the local conditional probabilities of all the nodes. The (JPD) of S is given as follows:

$$P(X_1,..., X_N) = \prod_{i=1}^{N} p(X_i / Pa(X_i))$$

A Dynamic Bayesian network (DBN) is a BN used to model a temporal stochastic process. It can be created by specifying network (structure and parameters) for two consecutive "time slices", and then "unrolling" it into a static network of the required size.

DBNs generalize two well-known signal modelling tools: Kalman filters for continuous state linear dynamic system (LDS) and Hidden Markov Models (HMMs) for classification of discrete state sequences. It has been shown that estimation in (LDSs) and inference in (HMMs) are special cases of inference in DBNs [11]. The focus of this paper is on a subclass of DBNs models called Switching Linear Systems or Switching Kalman Filter.

### 3.1 Switching Kalman Filter

Switching Kalman filter is a subclass of DBN and this type of network is useful for modeling piece-wise linear behavior (one way of approximating non-linear models), or multiple type or "mode" of behavior [11].

Consider a dynamical system whose parameters evolve in time according to some known model. This system can

be described using the following set of state-space equations:

$$X_t = A_t X_{t-1} + v_t$$

$$Y_t = C_t X_t + w_t$$

This model is called Linear Dynamical System (LDS), where $X_t \in \Re^N$ is the hidden state variable at time t, $Y_t \in \Re^M$ is the observation at time t, and $v_t \sim N(0,Q_t)$ and $w_t \sim N(0,R_t)$ are independent Gaussian noise. The parameters of model: $A_t, C_t, Q_t$ and $R_t$ are assumed to be time-invariant. Unfortunately, most systems are not linear and are subject to non-Gaussian noise. One approach to this problem and one we take in this paper, is to switch among *K* different linear models or take some linear combination of them (see Figure 5)[12].

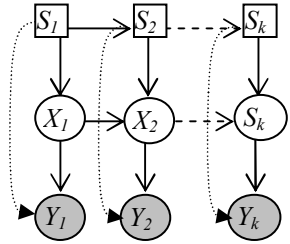

**Figure 5: A switching Kalman filter**

The variables $S_t$ are discrete and the variables $X_t$ and $Y_t$ are continuous. Other observations can be introduced in this model. The conditional probability distributions for this model are as follows:

*P ($X_t=x_t/X_{t-1}=x_{t-1}$, $S_t=i$) =N ($Ax_{t-1}+\mu_i$, Qi)*

*P ($Y_t=y/X_t=x_t$)=N ($Cx_t+\mu^Y$, R)*

*P ($S_t=j/S_{t-1}=i$) =B (i,j)*

An important issue in BN is the computation of posterior probabilities of variables given observations. Several researchers have been developed to compute the exact and/or the approximate inference algorithms for different distributions. The most commonly used to compute the exact inference algorithm for discrete Bayesian Networks is known as the JLO[1] algorithm [11]. The JLO algorithm is a recursive message passing algorithm that works on the junction tree of the BN. Namely, we need to find the posterior: *P ($X_t, S_t/ Y_t$)* in the case of switching Kalman filter.

## 4. Switching Kalman filter for road-matching

Road-matching involves applying a first filter which selects all the segments close to the estimated position of the vehicle. The goal is then to select the most likely segment(s) from this subset. Nowadays, since the geometry of roadmaps is more and more detailed, the number of segments representing roads is increasing. In the other hand, the map has an absolute error (10 meters) and relative error (1 meter). The road classification module is an important stage in the vehicle localization process because the robustness of the localization depends mainly on this stage. In order to take into account the error of several sensors or database used in this application, we introduce a concept which can manage multi-hypothesis in the formalism of BN.

---
[1] F.V. Jensen, S.L. Lauritzen et K.G. Olesen

### 4.1 BN model

For each selected segment $Carto_i$, we represent it by a Gaussian: $Carto_i \sim N (\mu_i, \sum_i)$. Where $\mu_i$ is the pose vector: $(x_i,y_i)$ is the projection of the estimated position on this segment and $\theta_i$ is the heading of the segment.

The proposed BN model for road-matching is illustrated in Figure 6. In this model we used two hidden variables. The discrete variable $S_k$ represents the segments of which the vehicle can be. The second is continuous variable; $X_k(x_i,y_i,\theta_i)$ represents the estimation of a vehicle for each segment candidate.

The graph represented in Figure 6 allows us to represent causal links between the variables. The variable $X_k$ is updated by observations $Carto_k$ and/or $GPS_k$ (if GPS is masked we use only the cartographical observations). This variable is multi-modal because it has been updated by the set of candidates segments ($Carto_k^{segi}$). The variable $S_k$ is update by cartographical observation ($Carto_k$) and the estimation is given by the hidden variable $X_k$.

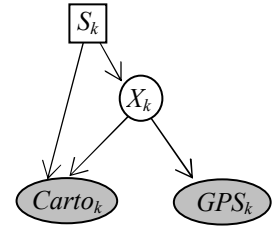

**Figure 6: Switching Kalman filter for a map-matching**

For each candidate segment one can build a cartographical observation given by projection of odometric estimation onto the segments. The cartographical observations and/or GPS observation are used to update variables $X_k$ and $S_k$. A result of Bayesian inference is a probability of each candidate segment. The synoptic of this algorithm is given by Figure 7.

Let us use a specific case study to illustrate the method. In Figure 8, the vehicle is traveling on the road represented by the segments 1 and 2. Estimation errors and digital map errors oblige the selection of the segment

3 in addition to the segment 2 to be treated also. So, two cartographical observations were generated in giving the same chance to each segment candidate. Consequently two estimations were generated also at the step t=k-1.

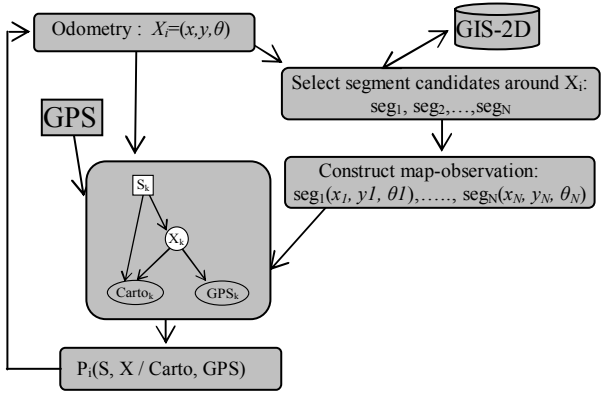

**Figure 7: Synoptic of road-matching using a BN**

At the instant k, three segments were selected all around the predicted pose. These segments were used to generate three observations. Then, using SKF, three estimations were provided. We plot only the most probable in green circle on the Figure 8.

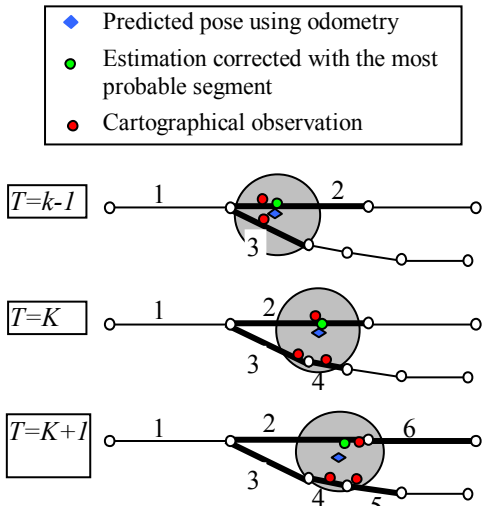

**Figure 8: Illustrative example of multi-hypotheses**

## 5. Experimental results

A test trajectory has been carried out at Compiègne in France with an experimental vehicle. The used GPS is a differential Trimble AgGPS132 receiver. For odometry, the ABS sensors is of the rear wheels of the vehicle are used.

The test trajectory is presented in Figure 9. In this experience, the GPS measurement was available in the beginning of the test trajectory. Then, the GPS still not used for 1.5Km. One can remark that in spite of the long GPS mask, the vehicle location is matched correctly. As matter of fact, the final estimated positions stay close to the GPS points. In Figure 9, we only presented the most probable SKF estimation of the pose.

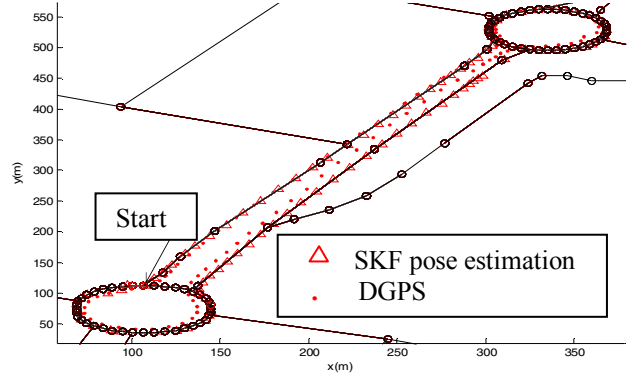

**Figure 9: pose estimation with SKF using odometry and cartographical observation (GPS was masked)**

In Figure 10, the estimation is plotted with a red ×. When the GPS measurement is not available, the method provides estimation for each segment candidate and the estimation of the most probable segment only is plotted. Then the GPS measurement it becomes available but the GPS position is then closer the segment which describe the other side of the road. In this case, the situation is an ambiguous parallel road situation. The method detects the ambiguity of this situation and selects all probable segments in this parallel road situation. The SKF manage all hypotheses until the elimination of the ambiguity. One can remark that, in spite the ambiguity, the road on which the vehicle is running in reality presents the highest probability. With these results, the proposed approach can manipulate and take into account the GPS error also.

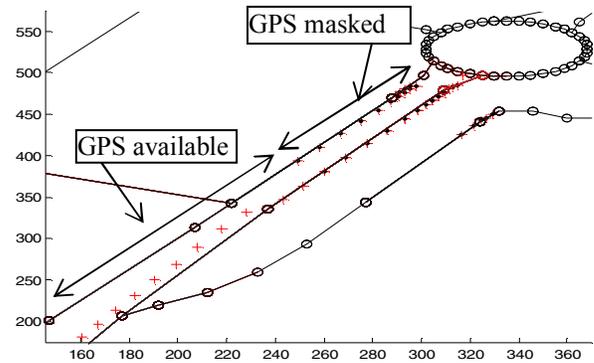

**Figure 10: Multi-hypothesis managed with SKF to treat parallel road situation**

In Figure 11, GPS was not available after the intersection. One can see that the method manage two hypotheses for seven steps then wrong hypothesis was eliminated. We can remark that, the good segment always presents the most important probability computed by the SKF inference.

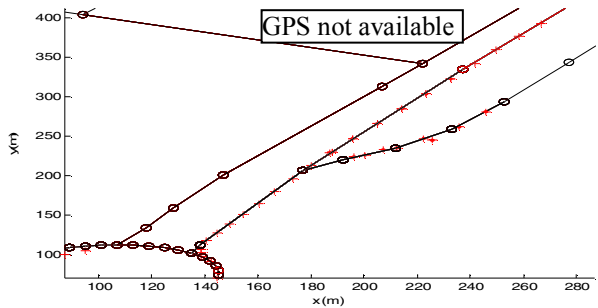

**Figure 11: Multi-hypothesis managed with SKF to treat junction road situation**

## 6. Conclusion

This article has presented a road-matching method based on a multi-sensor fusion approach. The main contributions of this work are the formalization of a road-matching method in the Switching Kalman filtering context and an experimental validation with real data.

An interesting characteristic of this approach is that it is flexible and modular in the sense that it can easily integrate other sensors. This feature is interesting because adding other sensors is a way to increase the robustness of the road-matching.

In this approach, the use of the digital map as an observation of the state space representation has been introduced. This observation is used in the Switching Kalman filter in the same way that the GPS data. It turned out in the experiments that the GPS measurements are not necessary available all the time, since the merging of odometry and roadmap data can provide a good estimation of the position over a substantial period. The strategy presented in this paper doesn't keep only the most likely segment. When approaching an intersection, several roads can be good candidates for this reason we manage several hypotheses until the situation becomes unambiguous.

## Acknowledgments

The authors wish to acknowledge the HEUDIASYC laboratory in the person of Philippe Bonnifait for his contribution.